\documentclass{article}
\usepackage{spconf,amsmath,graphicx}
\usepackage{subcaption}
\usepackage{amsmath,graphicx,amsfonts}
\usepackage{comment}
\usepackage{amssymb}
\usepackage{enumitem}

\def\auth{1pt}
\def\ftheta{\mathbf{f}_{\boldsymbol\theta}}

\name{Antoine Honoré, \hspace{\auth} Anubhab Ghosh, \hspace{\auth} Saikat Chatterjee\thanks{Correspondence: sach@kth.se
}}

\address{Digital Futures and KTH Royal Institute of Technology, Stockholm, Sweden\\ honore@kth.se, anubhabg@kth.se, sach@kth.se}

\title{Compressed Sensing of Generative Sparse-latent (GSL) Signals}

\begin{document}
\ninept
\maketitle

\begin{abstract}
We consider reconstruction of an ambient signal in a compressed sensing (CS) setup where the ambient signal has a neural network based generative model. The generative model has a sparse-latent input and we refer to the generated ambient signal as generative sparse-latent signal (GSL). The proposed sparsity inducing reconstruction algorithm is inherently non-convex, and we show that a gradient based search provides a good reconstruction performance. We evaluate our proposed algorithm using simulated data.
\end{abstract}
\begin{keywords}
Compressed sensing, generative models, inverse problems
\end{keywords}

\section{Introduction}

\noindent \textbf{Background:} Reconstruction of signals in an under-determined linear measurement setup, where the signals have generative models that are linear with latent sparse signals, has been thoroughly investigated in literature. This is primarily known as compressed sensing (CS) or compressed sampling \cite{Donoho_2006_Compressed_sensing, CS_introduction_Candes_Wakin_2008}.

In this article we refer to the linear measurement setup along-with the linear generative model as the traditional CS. 
Many practical algorithms exist for signal reconstruction using sparse-latent signals as a-priori knowledge. Algorithms are mainly divided as convex, greedy pursuits and Bayesian, and their suitable combinations \cite{Mallat_1994_Orthogonal_Matching_pursuits, Candes_Romberg_Tao_2006, Tropp_2007_OMP, Needell_2009_Regularized_OMP,Dai_2009_Subspace_pursuit, Chatterjee_Sundman_Vehkapera_Skoglund_TSP_2012, Candes_Wakin_Boyd_2008, Tipping_2001_RVM,Ji_Xue_Carin_2008_Bayesian_CS,Ambat_Chatterjee_Hari_2013_CoMACS}.
Convex algorithms, such as LASSO \cite{Tibshirani_1996_lasso}, use $\ell_1$-norm based penalties to promote sparsity. 

\noindent \textbf{Motivation:} In this article, we consider a CS setup where we have linear measurements of an ambient signal like traditional CS, but do not have the linear generative model for the ambient signal.
Instead, we assume non-linear, non-convex, generative models where an ambient signal is the output of a neural network excited by a sparse-latent input signal.
Signals which are generated in such a way are referred to as `\emph{generative sparse-latent}' (GSL) signals. 
Our objective is to design reconstruction algorithms for GSL signals in the CS setup. 

In the CS setup measuring GSL, we have a non-linear mapping between the measurement and the sparse-latent signal.
Due to the non-linear mapping, the reconstruction problem is inherently non-convex.
We design a reconstruction algorithm that can use a-priori knowledge of sparsity.
The reconstruction algorithm uses gradient search for sparse-latent estimation.
Using simulations we show that it is possible to achieve a good quality signal reconstruction, although the mapping between measurement vector and sparse-latent signal is highly non-linear.

Naturally a question arises: how to measure a degree of non-linearity? Such a measure could help us to perform simulation studies in a controlled manner.
To the best of authors' knowledge, we are not aware of any such measure.
Therefore we provide a simple measure based on how far a non-linear mapping is from a linear mapping.
This helps us to perform simulation studies in a controlled manner.

\noindent \textbf{Relevant literature:} In a CS setup, 
there exists prior work where neural networks-based generative models are used.
In \cite{boraCompressedSensingUsing2017a}, variational auto-encoders (VAEs) and generative adversarial networks (GANs) are used where the dimension of latent signal is (much) lower than the dimension of ambient signal.
In \cite{killedarSparsityDrivenLatent2021}, the authors considered GAN-based generative models and consider the case where the dimension of the latent is small.
Robustness of the CS for generative models was addressed in \cite{RobustCS_GenerativeModels_2020}.
The work \cite{weiDeepUnfoldingNormalizing2022} considered normalizing flows (NFs) \cite{klambauerSelfnormalizingNeuralNetworks2017} that are invertible neural networks. 
These works address CS of signals with generative models -- VAEs, GANs and NFs -- excited with dense, Gaussian latent signals.

\noindent \textbf{Our contributions:} Use of sparse-latent signals for generative models in CS setup is the major novelty of the article compared to previous works.
The contributions are: (a) Proposing the construction of GSLs.
(b) Defining a measure of non-linearity and using the measure to rank complex non-linear mapping functions to generate GSL signals.
(c) Proposing sparsity inducing reconstruction algorithms for CS of GSL signals and demonstrating their performances.

\section{Methods}

\subsection{GSL signals and its CS setup}\label{sec:GSL_signal}
A GSL signal $\mathbf{x}$ has the following generative model for the ambient signal:
\begin{eqnarray}
\mathbf{x} \triangleq \mathbf{f}_{\pmb{\theta}}(\mathbf{Bz}) \in \mathbb{R}^m,
\label{eq:GSL_Model}
\end{eqnarray}
where $\mathbf{z}\in \mathbb{R}^M$ is sparse, $\mathbf{B} \in \mathbb{R}^{m \times M}$ is a matrix, and $\mathbf{f}_{\pmb{\theta}}(.): \mathbb{R}^M \rightarrow \mathbb{R}^m$ is a neural network.
Neural networks have arbitrarily complex mapping functions and are differentiable.
We assume that $\| \mathbf{z} \|_0 \triangleq K < m \leq M$, where $\| . \|_0$ denotes $\ell_0$-norm.
The GSL model \eqref{eq:GSL_Model} is a generalization over the traditional linear generative sparse representation model where $\mathbf{f}_{\pmb{\theta}}(\cdot)$ is linear (or identity mapping).

In this article, we investigate the reconstruction of GSL signal $\mathbf{x}$ from linear CS measurements.
We assume that the GSL signal model and its parameters $\pmb{\theta},\mathbf{B}$ are known.
We do not consider the modeling issues and learning parameters of GSL signals.
In a CS setup, using linear measurements, we have $n$-dimensional noisy measurements:
\begin{eqnarray}
\mathbf{y} = \mathbf{Ax+n} = \mathbf{A}\mathbf{f}_{\pmb{\theta}}(\mathbf{Bz}) + \mathbf{n} \in \mathbb{R}^n, \,\, n<m,
\label{eq:GSL_CS}
\end{eqnarray}
where $\mathbf{x} \in \mathbb{R}^m$ is the signal to be reconstructed, $\mathbf{n}$ is additive measurement noise, and $n < m$.
Using the a-priori information $\mathbf{z}$ be sparse, we need to solve the under-determined setup.
We first estimate $\mathbf{z}$ as $\hat{\mathbf{z}}$ and then reconstruct $\mathbf{x}$ as $\hat{\mathbf{x}}=\mathbf{f}_{\pmb{\theta}}(\mathbf{B}\hat{\mathbf{z}})$.
Our investigation in this article is concentrated on designing practical algorithms for quality reconstructions.

\subsection{Optimization problems and reconstruction algorithm}
We first start with the GSL CS setup \eqref{eq:GSL_CS} where the measurement noise $\mathbf{n}$ is absent ($\mathbf{n=0}$).
In that case, a relevant optimization problem is
\begin{eqnarray}
\hat{\mathbf{z}} = \arg\min_{\mathbf{z}} \|\mathbf{z}\|_1 \,\, \mathrm{s.t.} \,\, \mathbf{y}=\mathbf{A}\mathbf{f}_{\pmb{\theta}}(\mathbf{Bz}).
\label{eq:P1_problem_GSL_noisefree}
\end{eqnarray}
The above problem has a resemblance with celebrated basis pursuit (BP) algorithm for the traditional CS.
The major point is that BP is convex (a linear program) while the optimization problem in \eqref{eq:P1_problem_GSL_noisefree} is non-convex.
BP provides exact reconstruction, i.e.  $\hat{\mathbf{z}} = \mathbf{z}$, under certain technical conditions.
In our understanding the optimization problem \eqref{eq:P1_problem_GSL_noisefree} is difficult to solve due to the (hard) equality constraint and presence of the non-linear mapping of the optimization variable $\mathbf{z}$.

While the measurement noise is absent or present, we address CS of GSL using the following optimization problem
\begin{eqnarray}
\hat{\mathbf{z}}=\arg\min_{\mathbf{z}}\left\{ \lambda_1\| \mathbf{y-A}\mathbf{f}_{\pmb{\theta}}(\mathbf{Bz})\|_2^2 + (1-\lambda_1) \|\mathbf{z}\|_1 \right\},
\label{eq:P2_problem_GSL}
\end{eqnarray}
where $\lambda_1 \in \left[0,1\right]$ is a regularization parameter.
The above optimization problem bears resemblance to the celebrated LASSO or basis pursuit denoising (BPDN) \cite{Chen_Donoho_Saunders_1998}.
While LASSO / BPDN is convex, the optimization problem \eqref{eq:P2_problem_GSL} is non-convex.
To solve the problem we use gradient search.
The gradient search is
\begin{eqnarray} \label{eq:GDscheme}
\hat{\mathbf{z}}_{k+1} = \hat{\mathbf{z}}_k - \eta \left. \frac{\partial \mathcal{L}}{\partial \mathbf{z}} \right|_{\mathbf{z}=\hat{\mathbf{z}}_k},
\end{eqnarray} 
where $\mathcal{L}(\mathbf{z})=\lambda_1\| \mathbf{y-A}\mathbf{f}_{\pmb{\theta}}(\mathbf{Bz})\|_2^2 + (1-\lambda_1) \|\mathbf{z}\|_1$ and $\eta$ is a tunable learning rate parameter.
We use ADAM optimizer \cite{kingmaAdamMethodStochastic2017} to realize the gradient search.
It is expected that the success of gradient search will be highly dependent on the degree of non-linearity in $\mathbf{f}_{\pmb{\theta}}(.)$, and the choices of hyper-parameters $\lambda_1$ and $\eta$.

On quest of theoretical analysis, several questions can arise. Example questions are: (a) How good is the gradient search for solving the optimization algorithm \eqref{eq:P2_problem_GSL}? (b) Can we reach to a globally optimum point? (c) If not, then how far is the local optimum point from the global optimum point?
These questions are non-trivial given the optimization problem is non-convex.
We do not deliberate on such theoretical questions in this article.
Instead we investigate whether we can achieve empirical success using simulations.

\subsection{Competing optimization problems}
Competing optimization problems may use $\ell_2$-norm based penalties on the ambient signal $\mathbf{x}$ or the latent signal $\mathbf{z}$.
Without any signal structure imposition, a standard optimization problem is 
\begin{eqnarray}
\min_{\mathbf{x}} \left\{ \kappa \| \mathbf{y-A} \mathbf{x} \|_2^2 + (1 - \kappa) \|\mathbf{x}\|_2^2 \right\},
\label{eq:regularized_least_squares}
\end{eqnarray}
where $\kappa \in \left[0,1\right]$ is a tunable parameter.
Note that the optimization problem is convex (it is a regularized least-squares).

To use signal structures, algorithms in \cite{boraCompressedSensingUsing2017a, weiDeepUnfoldingNormalizing2022} use the following non-convex optimization problem: \begin{eqnarray}
\min_{\mathbf{z}} \left\{ \lambda_2 \| \mathbf{y-A} \mathbf{f}_{\pmb{\theta}}(\mathbf{B}\mathbf{z}) \|_2^2 + \left(1 - \lambda_2\right) \|\mathbf{z}\|_2^2 \right\},
\label{eq:P2_problem_GM}
\end{eqnarray}
where $\lambda_2 \in \left[0,1\right]$ is a regularization parameter.
In \cite{boraCompressedSensingUsing2017a}, authors investigated use of GAN and decoders of auto-encoders as the generative mapping function $\mathbf{f}_{\pmb{\theta}}(\mathbf{B}\mathbf{z})$, where $\mathrm{dim}(\mathbf{z}) \ll \mathrm{dim}(\mathbf{x})$ and it can happen that $\mathrm{dim}(\mathbf{z}) < \mathrm{dim}(\mathbf{y})$.
Therefore the problem \eqref{eq:P2_problem_GM} while non-convex, may not be fully under-sampled.
On the other hand, the work of \cite{weiDeepUnfoldingNormalizing2022} used normalizing flows (NFs) as the generative mapping function, where $\mathrm{dim}(\mathbf{z}) = \mathrm{dim}(\mathbf{x})$ due to invertibility requirement.
Therefore $\mathrm{dim}(\mathbf{z}) > \mathrm{dim}(\mathbf{y})$ and the problem \eqref{eq:P2_problem_GM} is under-sampled in \cite{weiDeepUnfoldingNormalizing2022}.
The optimization problem \eqref{eq:P2_problem_GM} can be considered state-of-the-art in CS for generative models.

\subsection{A measure of non-linearity}
\label{subsec:measure_of_non-linearity}

In this subsection, we deliberate on how to measure non-linearity.
We conjecture that the optimization problem \eqref{eq:P2_problem_GSL} becomes hard to solve if $\mathbf{f}_{\pmb{\theta}}$ becomes highly non-linear.
This in turns raises the question: Does the optimization problem gets harder if $\mathbf{f}_{\pmb{\theta}}$ gets more non-linear according to some metric of non-linearity? Then a natural queries can be: How to measure non-linearity? How to think that a function is more non-linear that another function? Can we conduct a systematic study to show that solving optimization problem \eqref{eq:P2_problem_GSL} becomes harder with the increase in some measure of non-linearity?

We are not aware of a universal measure of non-linearity.
Therefore we define a measure from a first principle - how much a non-linear mapping is away from an optimal linear mapping.
The procedure is as follows.

For a given $\mathbf{f}_{\pmb{\theta}}$, we first construct a dataset $\{(\mathbf{x}_j,\mathbf{z}_j)\}_{j=1}^J$ comprised of $J$ data samples, where $(\mathbf{x}_j,\mathbf{z}_j)$ is $j$'th datum.
We use the generative procedure $\mathbf{x}=\ftheta(\mathbf{B}\mathbf{z})$ where $\mathbf{z}$ need not be sparse; $\mathbf{z}$ can be dense.
Using this dataset, we construct the linear minimum-mean-square-estimation (LMMSE) as
\begin{equation}
    \hat{\mathbf{x}}_L = C_{\mathbf{x}\mathbf{z}}(C_{\mathbf{zz}} + \lambda I)^{-1}(\mathbf{z} - \mu_{\mathbf{z}}) + \mu_{\mathbf{x}},
\end{equation}
where $C_{\mathbf{x}\mathbf{z}}$ is the empirical cross-covariance between $\mathbf{x}$ and $\mathbf{z}$, $C_{\mathbf{zz}}$ is the empirical covariance of $\mathbf{z}$, and $\mu_\mathbf{z}$ and $\mu_\mathbf{x}$ are the empirical means of $\mathbf{z}$ and $\mathbf{x}$, respectively, and $\lambda > 0$ is a tunable regularization parameter.
The strength of non-linearities is quantified with a metric referred to as normalized-non-linearity-measure (NNLM) and defined as follows:
\begin{eqnarray}
\begin{array}{rl}
\mathrm{NNLM}(\ftheta(.)) & =\frac{\mathbb{E}[\|\mathbf{x}-\hat{\mathbf{x}}_L\|_2^2]}{\mathbb{E}[\|\mathbf{x}\|^2_2]} = \frac{\mathbb{E}[\|\ftheta(\mathbf{B}\mathbf{z})-\hat{\mathbf{x}}_L\|_2^2]}{\mathbb{E}\left[\Vert \ftheta(\mathbf{B}\mathbf{z})\Vert_2^2\right]}.
\end{array}
\end{eqnarray}
A high value of NNLM translates to a high non-linearity, and this should be associated with larger reconstruction errors.
We choose $\lambda$ appropriately using a cross-validation approach.
\begin{figure}[t]
    \centering
    \includegraphics[width=0.5\textwidth]{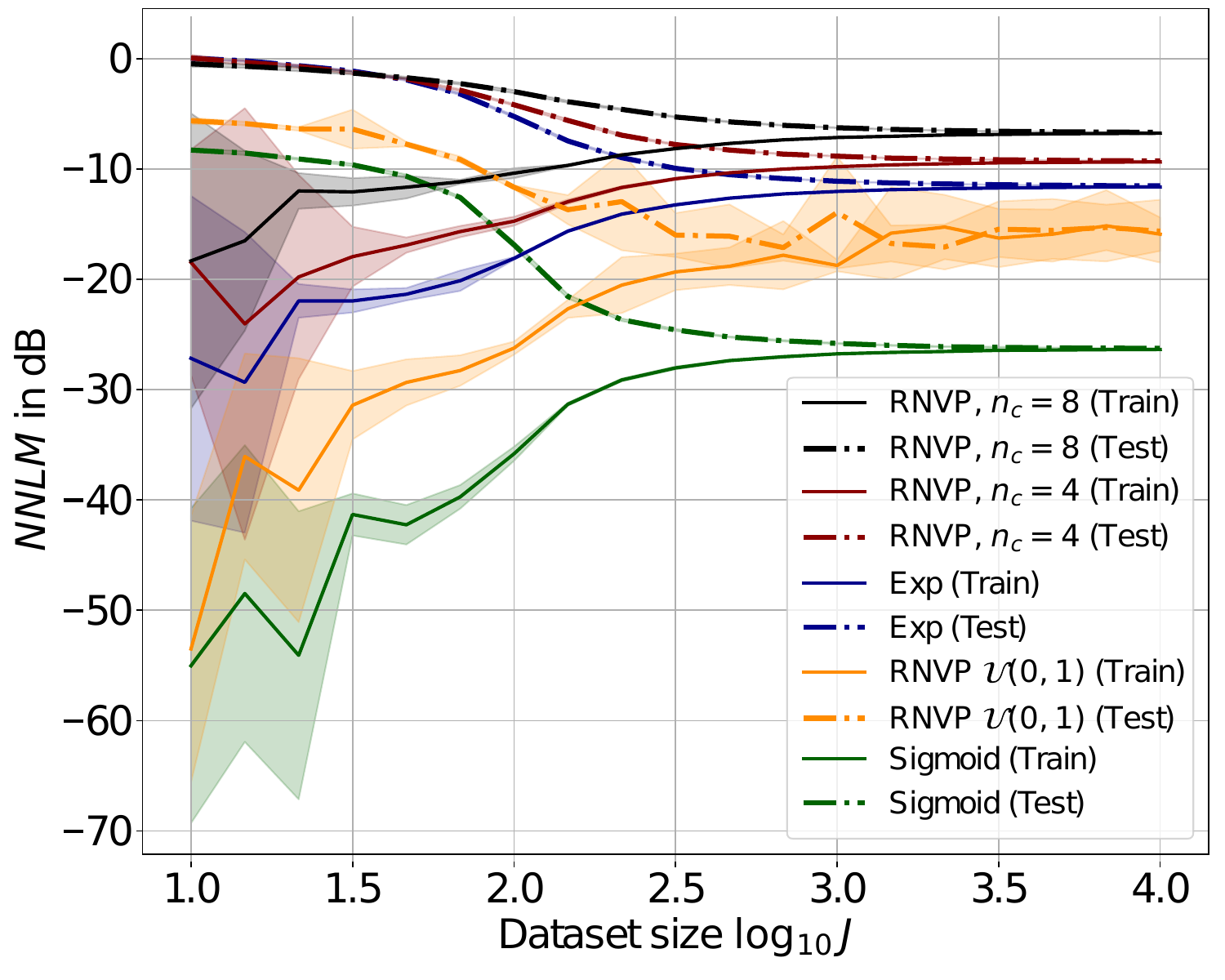}
    \caption{Analysis of non-linearity}
    \label{fig:nonlinearity}
\end{figure}

\section{Experiments}
In this section we perform simulations using synthetic data.
This helps our investigation in a controlled manner.
The simulation were carried out in Python\footnote{The code is available upon request.} using the PyTorch toolbox \cite{paszkePyTorchImperativeStyle2019}.
The experiments are carried out as follows.
\begin{enumerate}[nolistsep, noitemsep]
    \item In our first study, we use pretrained and untrained $\ftheta$ and study their NNLM score.
    This helps us to visualize how their degree of non-linearity varies.
    \item The second study compares reconstruction performance of three competing algorithms for GSL CS, and show that the proposed algorithm \eqref{eq:P2_problem_GSL} performs the best.
\end{enumerate}

\begin{figure}[t]
    \centering
    \includegraphics[width=0.5\textwidth]{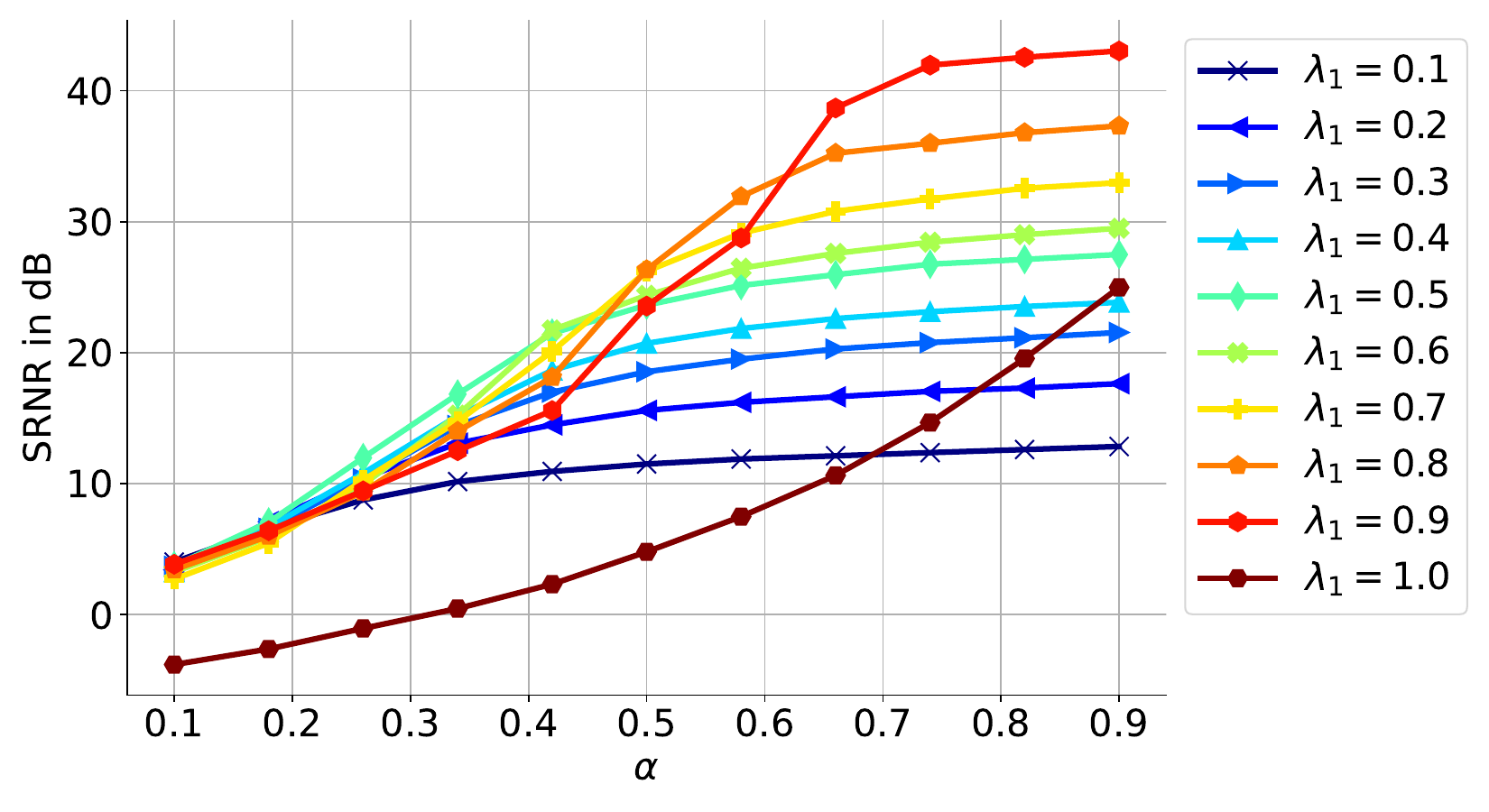}
    \caption{
    SRNR performance of RNVP ($n_c=4$) versus the sub-sampling ratio ($\alpha$) for different values of $\lambda_1$.
    }
    \label{fig:srnr_lamb_choice}
\end{figure}

\subsection{First study - On non-linearity}
\label{subsec:On_non-linearity}
\begin{figure*}[t]
    \centering
    \includegraphics[width=1.02\textwidth]{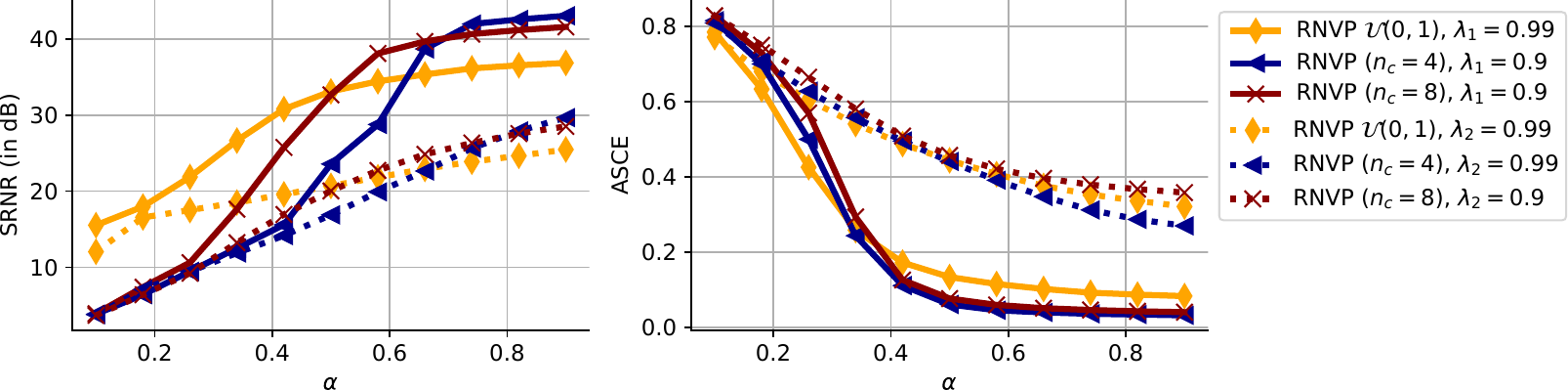}
    \caption{Comparison of reconstruction methods in terms of SRNR and ASCE for $3$ non-linearities.}
    \label{fig:srnr_asce_comparison}
\end{figure*}
In this study we consider several non-linear functions. 
(a) One-layer feedforward neural networks with either sigmoid or exponential activations. 
(b) RealNVP NF \cite{dinhDensityEstimationUsing2016a} with various layers.
We conjecture that as the number of layers increases the mapping function becomes more non-linear. 
RealNVP (henceforth referred to as RNVP) is known to be a powerful generative model (NF), recently used in \cite{weiDeepUnfoldingNormalizing2022}, and hence we choose it for our study. We use various configurations of RNVP.

In one configuration, a RNVP with 16 coupling layers is pre-trained to map a isotropic Gaussian latent source to a uniform ambient distribution.
The trained model is later used for our purpose where a GSL signal is generated by exciting the trained model using a sparse-latent signal.
In other configurations we use untrained RNVP with 4 coupling layers and 8 coupling layers.
In order to guarantee that the randomly initialized models are non-linear, the activation functions and the parameters priors for $\pmb{\theta}$ are chosen according to \cite{klambauerSelfnormalizingNeuralNetworks2017}.
Finally, the components of $\mathbf{B}$ are drawn from a normal distribution and the column are $\ell_2$-normalized.

In the experiments, we use $100$-dimensional isotropic Gaussian $\mathbf{z}$ to create signal $\mathbf{x}$, as mentioned in section \ref{sec:GSL_signal}.
Linear estimators in high dimension are subject to the curse of dimensionality, making it possible to accurately estimate non-linear data with a linear transform on the training data set.
We thus use the NNLM measure with a varying number of data points $J$, and by computing the NNLM on the training and a held-out test set.

The experimental results are shown in Fig. \ref{fig:nonlinearity} for training data (used to learn the LMMSE parameters) and held out test data vis-a-vis number of data points $J$.
In the figure, a higher NNLM value means a higher amount of non-linearity.
From Fig. \ref{fig:nonlinearity} we can observe that increase in the number of coupling layers from 4 (denoted as RNVP, $n_c=4$) to 8 (denoted as RNVP, $n_c=8$) for RNVP with random parameters leads to an increase in NNLM, which indicates a higher amount of non-linearity.
The trained RNVP with 16 layers shows a lower NNLM than the RNVPs with random parameters, since it effectively learns the mapping of the Gaussian multivariate cumulative distribution function (by probability integral transform \cite{angus1994probability}).
The trend of the curves is typical for training-testing scenarios.
The ranking of the RNVP models according to an increasing trend of non-linearity is as follows: (i) RNVP $\mathcal{U}(0,1)$ as the pretrained RNVP, (ii) RNVP with random parameters and 4 coupling layers, (iii) RNVP with random parameters and 8 coupling layers.

\subsection{Second study - Performance and comparison}
For GSL CS, we now study the performances for three RNVP models: RNVP $\mathcal{U}(0,1)$,  RNVP $n_c=4$, and RNVP $n_c=8$.
The GSL signals are produced using sparse-latent $\mathbf{z}$. 

We choose $m=M=100$ and a sparsity level of $10\%$, i.e.  $K=\|\mathbf{z}\|_0=0.1 M=10$.
The support-set of $\mathbf{z}$ is chosen randomly and the non-zero components of $\mathbf{z}$ are drawn from a standard Gaussian.
We add additive white Gaussian noise $\mathbf{n}$ in the signal-to-noise-ratio level 30 dB.
We introduce sub-sampling ratio $\alpha = \frac{n}{m} \in[0.1,0.9]$. We vary $\alpha$ and study reconstruction performance for GSL CS.

Following \cite{Chatterjee_Sundman_Vehkapera_Skoglund_TSP_2012}, we use two performance measures for the evaluation of the reconstruction: (1) signal to reconstruction noise ratio (SRNR) $=10\log_{10}\frac{\mathbb{E}[\|\mathbf{x}\|_2^2]}{\mathbb{E}[\|\mathbf{x}-\hat{\mathbf{x}}\|_2^2]}$, and (2) average support cardinality error (ASCE) $= 1-\frac{1}{K}\mathbb{E}\lbrace |\mathcal{S}_\mathbf{z} \cap {\mathcal{S}}_{\hat{\mathbf{z}}}|\rbrace$, where $\mathcal{S}_\mathbf{z}$ denotes the $K$-size support of $\mathbf{z}$.
A $K$-size support set comprises of the indices of the $K$ components of largest amplitude of $\mathbf{z}$. 

We first study the effect of tuning $\lambda_1$ of \eqref{eq:P2_problem_GSL}.
Fig. \ref{fig:srnr_lamb_choice} shows SRNR performance of RNVP, $n_c=4$, against $\alpha$ for different $\lambda_1$.
We note that increase in $\lambda_1$ helps reconstruction and there exists a suitable range for an appropriate choice of $\lambda_1$.
Note that $\lambda_1=1$ leads to no sparsity promoting penalty term and the performance degrades badly.
The experiment confirms the importance of the sparsity-promoting penalty $\| \mathbf{z} \|_1$, and suggests that it is necessary to choose a suitable $\lambda_1$ to achieve good results.

Next we study the performances of three RNVP models and compare them.
We include $\ell_2$-penalty based optimization algorithm \eqref{eq:P2_problem_GM} where $\lambda_2$ is chosen appropriately. Fig. \ref{fig:srnr_asce_comparison} shows SRNR and ASCE performances of methods.
In the figure, $\lambda_1$ and $\lambda_2$ denotes the use of the optimization problem \eqref{eq:P2_problem_GSL} and \eqref{eq:P2_problem_GM}, respectively.
It is clear that the use of sparsity promoting penalty $\| \mathbf{z} \|_1$ helps to achieve better reconstruction performances than the $\| \mathbf{z} \|_2^2$ penalty. 

Another interesting result is that reconstruction algorithms show success according to the non-linearity level of $\ftheta$.
Following our remarks at the end of section \ref{subsec:On_non-linearity}, we find that the performance curves in Fig. \ref{fig:srnr_asce_comparison} are consistent with non-linearity level at the lower range of $\alpha$.
Lower range of $\alpha$ is of high importance in CS.

We also investigated the regularized least-squares \eqref{eq:regularized_least_squares} and found that the performance are not promising.
We skip those results to show in this article due to brevity.

\begin{figure}[ht]
    \centering
    \includegraphics[width=0.5\textwidth]{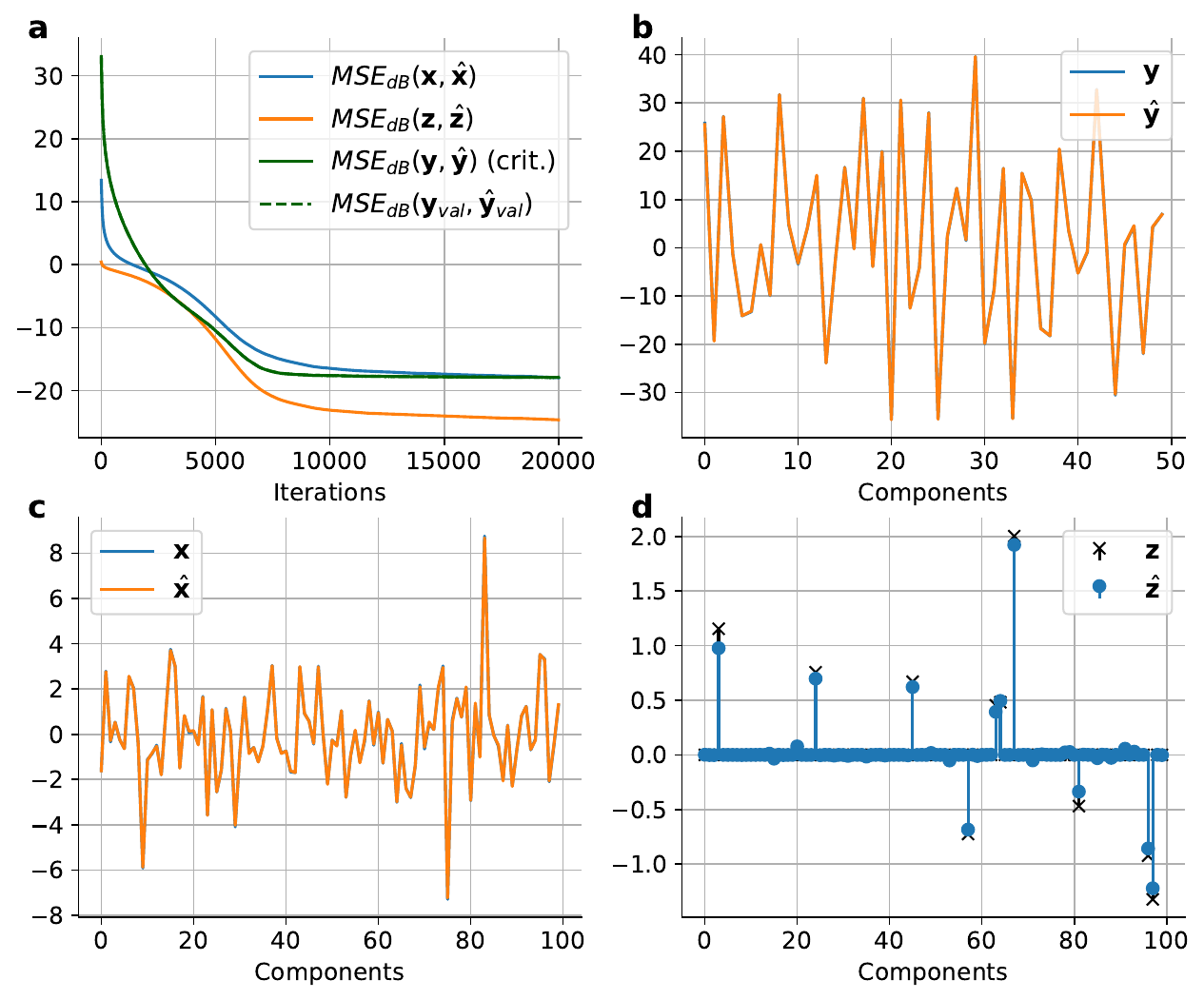}
    \caption{Example of a successful reconstruction for one the $J$ datum generated with a RNVP ($n_c=4$) model and using $\ell_1$ regularization with $\lambda_1=0.9$ and $\alpha=0.5$.}
    \label{fig:example_reconstruction}
\end{figure}

Finally we show a GSL signal realization using RNVP, $n_c=4$, and its reconstruction using the optimization algorithm \eqref{eq:P2_problem_GSL}. Fig. \ref{fig:example_reconstruction}c shows a realization of GSL signal $\mathbf{x}$ and Fig. \ref{fig:example_reconstruction}d shows its corresponding sparse-latent signal $\mathbf{z}$.
This visualization helps to show that it is possible to generate ambient signals using sparse-latents.
We also show reconstructed $\hat{\mathbf{x}}$ and $\hat{\mathbf{z}}$. Fig. \ref{fig:example_reconstruction}a demonstrates the gradient descent scheme in \eqref{eq:GDscheme} and shows that the algorithm indeed converges to a local optimum.
For the sake of completeness, we also show in Fig. \ref{fig:example_reconstruction}b, the observation signal $\mathbf{y}$ and its reconstruction $\hat{\mathbf{y}}$ using $\hat{\mathbf{x}}$. 

\subsection{Scope of improvements and challenges}
While we mainly concentrate on the optimization problem \eqref{eq:P2_problem_GSL} and its solution in this article using gradient descent, we delineate further improvement scopes and challenges.

Our first approach could be iteratively reweighted $\ell_1$ minimization, where we solve
\begin{eqnarray}
\hat{\mathbf{z}}=\arg\min_{\mathbf{z}} \left\{ \beta \| \mathbf{y-A}\mathbf{f}_{\pmb{\theta}}(\mathbf{Bz})\|_2^2 + (1 - \beta) \|\mathbf{Wz}\|_1 \right\},
\end{eqnarray}
where $\beta$ is a tunable parameter and $\beta \in \left[0,1\right]$.
The motivation of the improvement is from the work \cite{Candes_Wakin_Boyd_2008} applied to the traditional CS.
Here $\mathbf{W}$ can be iteratively chosen to enhance sparsity in the same way of \cite{Candes_Wakin_Boyd_2008}.
In the current iteration, we use $\mathbf{W}$ constructed from the solution of previous iteration.
The same gradient-descent based approach of solving \eqref{eq:P2_problem_GSL} can be used here in each iteration.

The next approach could follow the idea of iterative-reweighted-least-squares (IRLS) algorithm that has been in use for solving the traditional CS.
In that case we write $\|\mathbf{z}\|_1$ in terms of $\ell_2$-norm based representation, as $\mathbf{z}^{\top} [\mathrm{diag}(\mathbf{1}./ |\mathbf{z}|)]\mathbf{z}$, where $\mathbf{1}./ |\mathbf{z}| = \left[\frac{1}{|z_1|} \, \frac{1}{|z_2|} \,\hdots \frac{1}{|z_M|} \,\right]$.
In the current iteration we can use the solution of $\mathbf{z}$ from previous iteration to construct $\mathbf{1}./ |\mathbf{z}|$.
In each iteration, we can use the same approach of solving \eqref{eq:P2_problem_GM}.
Note that we did not experiment with the two possible scopes of improvements discussed above.

Finally, let us consider a Bayesian probabilistic approach.
In that case, we wish to (1) find a posterior $p(\mathbf{z}|\mathbf{y})$ using a sparse prior $p(\mathbf{z})$, e.g. Laplacian prior or Gaussian-Gamma prior, and (2) find the posterior $p(\mathbf{x}|\mathbf{y})$ (rather than a point estimate).
While relevance-vector-machine (RVM) and Bayesian CS with Gaussian-Gamma prior are used for the traditional CS \cite{Tipping_2001_RVM, He_Carin_2009_StructuredBayesian_CS}, an extension of such works for \eqref{eq:GSL_CS} perhaps comes with technical challenges.
We believe the technical challenge will arise due to complicated non-linear functions $\mathbf{f}_{\pmb{\theta}}$.
This will have repercussions to compute $p(\mathbf{z}|\mathbf{y})$ and then translate the result to compute $p(\mathbf{x}|\mathbf{y})$.  

\section{Conclusion}
In this study we introduced GSL signals and address their reconstruction in CS setup.
We propose a measure of non-linearity for the generative models.
We show that our measure is relevant to determine the quality of reconstruction at a low number of measurements.
The proposed sparsity inducing reconstruction algorithm is shown to outperform competing methods in terms of the chosen performance measures -- SRNR and ASCE.
Future works include theoretical analysis and designing more efficient algorithms; also learning the parameters of GSL signal models for real signals.

\section{Acknowledgement}The authors thank Borja Rodríguez Gálvez (KTH) for fruitful discussions regarding the design of the non-linearity metric.

\bibliographystyle{IEEEbib}
\bibliography{2022sparse_reconstruction,biblio_saikat_Pub,biblio_saikat_CS1}

\end{document}